\documentclass[10pt,twocolumn,letterpaper]{article}

\usepackage{cvpr}
\usepackage{times}
\usepackage{epsfig}
\usepackage{graphicx}
\usepackage{amsmath}
\usepackage{amssymb}

\usepackage[pagebackref=true,breaklinks=true,letterpaper=true,colorlinks,bookmarks=false]{hyperref}

\cvprfinalcopy 

\def\cvprPaperID{***} 

\ifcvprfinal\fi
\begin{document}

\title{Deep End2End Voxel2Voxel Prediction}

\author{Du Tran$^{1,2}$, Lubomir Bourdev$^3$, Rob Fergus$^1$, Lorenzo Torresani$^2$, Manohar Paluri$^1$\\
$^1$Facebook AI Research, $^2$Dartmouth College, $^3$UC Berkeley\\
{\tt\small \{dutran,lorenzo\}@cs.dartmouth.edu} \ \ \ {\tt\small \{robfergus,mano\}@fb.com} \ \ \ {\tt\small lubomir.bourdev@gmail.com}
}

\maketitle

\begin{abstract}
%

Over the last few years deep learning methods have emerged as one of the most prominent approaches for video analysis. However, so far their most successful applications have been in the area of video classification and detection, i.e., problems involving the prediction of a single class label or a handful of output variables per video.  Furthermore, while deep networks are commonly recognized as the best models to use in these domains, there is a widespread perception that in order to yield successful results they often require time-consuming architecture search, manual tweaking of parameters and computationally intensive pre-processing or post-processing methods.

In this paper we challenge these views by presenting a deep 3D convolutional architecture trained end to end to perform voxel-level prediction, i.e., to output a variable at every voxel of the video. Most importantly, we show that the same exact architecture can be used to achieve competitive results on three widely different voxel-prediction tasks: video semantic segmentation, optical flow estimation, and video coloring. The three networks learned on these problems are trained from raw video without any form of preprocessing and their outputs do not require post-processing to achieve outstanding performance. Thus, they offer an efficient alternative to traditional and much more computationally expensive methods in these video domains.
 \end{abstract}

\section{Introduction}

\begin{figure}[t]
\begin{center}
   \includegraphics[width=0.8\linewidth]{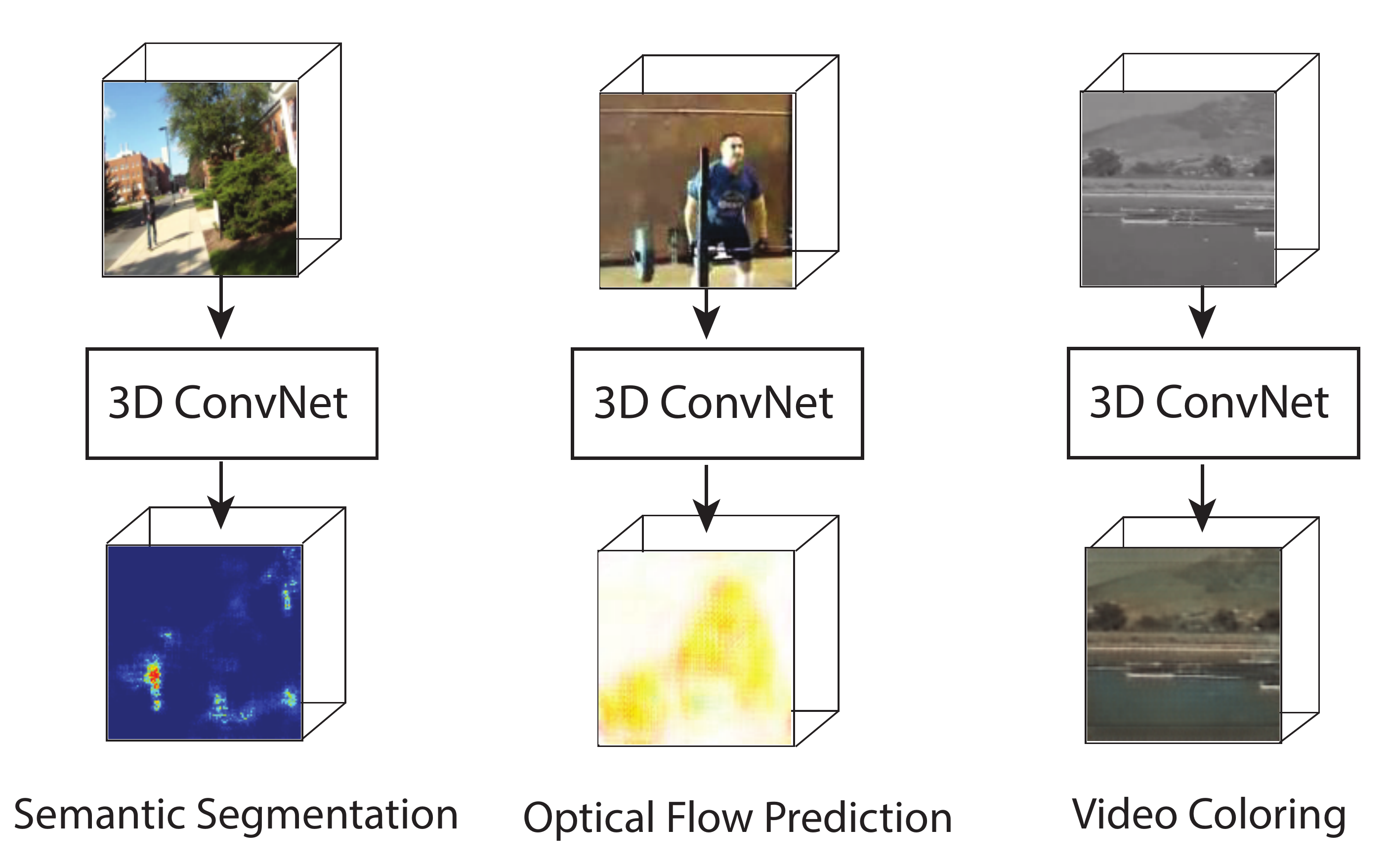}
\end{center}
\vspace{-12pt}
   \caption{{\bf Voxel to voxel prediction}: is a fine-grained video understanding task where the algorithm need to infer a variable for each input voxel. The problem has many potential applications including video semantic segmentation, optical flow prediction, depth estimation, and video coloring.}
\label{fig:intro}
\end{figure}

During the last decade we have witnessed a tremendous growth in the number of videos created and shared on the Internet thanks to the advances in network bandwidth and computation. In turn this has lead to a strong  effort toward the creation of better tools and apps to search, browse and navigate this large and continuously expanding video collections. This poses new challenges for the computer vision community and gives new motivations to build better, faster and more generally applicable video analysis methods.

In the still-image domain deep learning has revolutionized the traditional computer vision pipeline, which typically consisted of: pre-processing, hand-construction of visual features, training of a learning model, and post-processing. Instead, the successful introduction of deep convolutional neural network~\cite{Krizhevsky12,girshick13,overfeat,SimonyanZ14a} has shown that much better results can be obtained through end to end learning on very large collections of image examples, where the network is trained on raw image input and it directly predicts the target output. Besides the demonstrated advantages in improved accuracy, these end to end learned models have also been shown to be often more computationally efficient than traditional hand-designed approaches because they eliminate the need for computationally expensive pre-processing and post-processing steps and because convolution can run very fast, particularly on GPUs.

The video domain is also harnessing the benefits of this revolution but it is still lagging compared to the image setting~\cite{Donahue13,Zhou2014,Googlenet}. In particular, most of the end to end learning approaches for video analysis have been introduced in the area of classification and detection ~\cite{Karpathy14,SimonyanZ14,Wang2013IJCV,TranBFTP15} and involve predicting a single label or few output variables per video. However, there are many computer vision problems that require labeling every single voxel of a video. Examples include optical flow computation, video semantic segmentation, depth estimation and video coloring. There have been only a few attempts at approaching these pixel-labeling problems with deep learning~\cite{LongSD15,FlowNet,EigenNIPS14} for images. One of the reasons is that deep networks typically involve a large set of pooling layers which significantly lower the spatial resolution of the output. In order to output pixel labels at the original resolution, several ``unpooling'' strategies have been proposed, including simple upsampling, and multi-scale approaches. One of the most promising solution in this genre is learning convolution filters that upsample the signal. The primary benefit of convolutional upsampling is that it only requires  learning a small number of location-agnostic filters and thus it can be carried out with limited training data. 

The objective of our work is to demonstrate that 3D convolutional networks (3D ConvNets) with upsampling layers enable highly effective end to end learning of voxel to voxel prediction models on various video analysis problems. Instead of building a highly specialized network for each problem, our goal is to show that the same 3D ConvNet architecture trained on three distinct application domains (optical flow prediction, semantic segmentation, video coloring) can produce competitive results on each of them. Although a thorough architecture search is likely to yield improved results, we find it useful to employ a single network model for the three distinct tasks to convey the message that deep learning methods do not necessarily require to be highly specialized for the task at hand in order to produce good results. For the same reason, we do not employ any pre-processing or post-processing of the data. Because our model is fully convolutional, it involves a small number of learning parameters which can be optimized with limited amount of supervised data. Furthermore, the elimination of computationally expensive pre-processing and post-processing methods (such as CRF optimization or variational inference) and the exclusive reliance on efficient convolution implies that our learned models run very fast and can be used in real-time video-processing applications such as those arising in big-data domains. 


In summary, our work provides the following findings:
\begin{enumerate}
\item  Fully convolutional 3D ConvNets enable end to end learning of voxel to voxel prediction models with limited training data.
\item The same exact architecture can be employed to obtain competitive results on three different voxel-labeling applications: optical flow estimation, semantic segmentation of image sequences, and video coloring.
\item In domains where supervised training data is scarce (such as in the case of optical flow), we can train our end to end learning model on the output of an existing hand-designed algorithm. We show that this results in a 3D ConvNet that achieves slightly better accuracy than the complex hand-tuned vision method but, most importantly, it is significantly more efficient.
\item While fine-tuning a pre-trained model helps in most cases, it actually hurts when the new domain requires visual features that are quite distinct from those of the pre-learned model, such as in the case of fine-tuning an action recognition network for optical flow estimation.
\end{enumerate}

\section{Related Work}



Video analysis has been studied by the computer vision community for decades. Different approaches were proposed for action recognition including: tracking-based methods~\cite{Efros03}, bag-of-visual words~\cite{FeiFei07}, biologically-inspired models~\cite{Jhuang07}, space-time shapes~\cite{Irani05}, HMMs~\cite{Ikizler08}, and template-based Action-Bank~\cite{ActionBank}. Different spatio-temporal features were also introduced for video and action classification: Spatio-Temporal Interest Points~\cite{Laptev03}, improved Dense Trajectories~\cite{Wang2013IJCV}. Various methods were used for action and video event detection~\cite{HJSeo09,LCao10,Yuan11}. Although these methods showed to work reasonably well, they are not scalable because most of them require computational intensive steps during preprocessing (e.g. tracking, background subtraction, or feature extraction) or post-processing (CRF, variational inference).

Deep learning methods have recently shown good on different computer vision problems~\cite{Googlenet,overfeat,Ng15,girshick13,Bertasius15}. Thanks to their large learning capacity and the ability to optimize all parameters end to end, these methods achieved good performance on classification~\cite{Krizhevsky12} and feature learning~\cite{Googlenet,TranBFTP15} provided that there is sufficient supervised training data. Among the deep learning approaches, our proposed method is most closely related to the depth estimation method described in~\cite{EigenNIPS14}, the Fully Convolutional Network (FCN)~\cite{LongSD15}, and FlowNet~\cite{FlowNet}. Our method shares with these approaches the property of making pixel-level predictions. However, all these prior methods are designed for still image problems, while our method operates on videos. To the best of our knowledge, our method is the first one addressing end-to-end training of video voxel prediction.

\section{Video Voxel Prediction}
\label{sec:formulation}

{\bf Problem statement}. The input to our system is video with size $C \times L \times H \times W$, where $C$ is the number of color channels, $L$ is its temporal length (in number of frames), and $H, W$ are the frame height and width. Then, a voxel prediction problem requires producing a target output of size $K \times L \times H \times W$, where $K$ is an application-dependent integer denoting the number of output variables that need to be predicted {\em per voxel}. It is worth nothing that the size of the input video and the output prediction are the same, except only for the number of input channels $C$ and the number of output channels $K$ are different. Normally, $C=3$ for the case of color video inputs and $C=1$ for gray-scale inputs. For the three voxel-prediction applications considered in this paper, $K$ will have the following values: $K=2$ for optical flow estimation (the horizontal and vertical motion displacement for each voxel), $K=3$ for video coloring (the three color channels) and $K$ will be equal to the number of semantic classes in the case of video semantic segmentation.   

{\bf Proposed approach}. We propose a novel and unified approach for video voxel prediction based on a 3D ConvNet architecture with 3D deconvolution layers. We show the generality of the model by demonstrating that a simple unified architecture can work reasonably well across different tasks without any engineering efforts in architecture search. Since our method uses 3D deconvolution layers, we will start by briefly explaining the idea of 2D deconvolution~\cite{ZeilerF14,LongSD15} and then present our architecture based on 3D deconvolution for voxel prediction.

\begin{figure}
\begin{center}
   \includegraphics[width=0.96\linewidth]{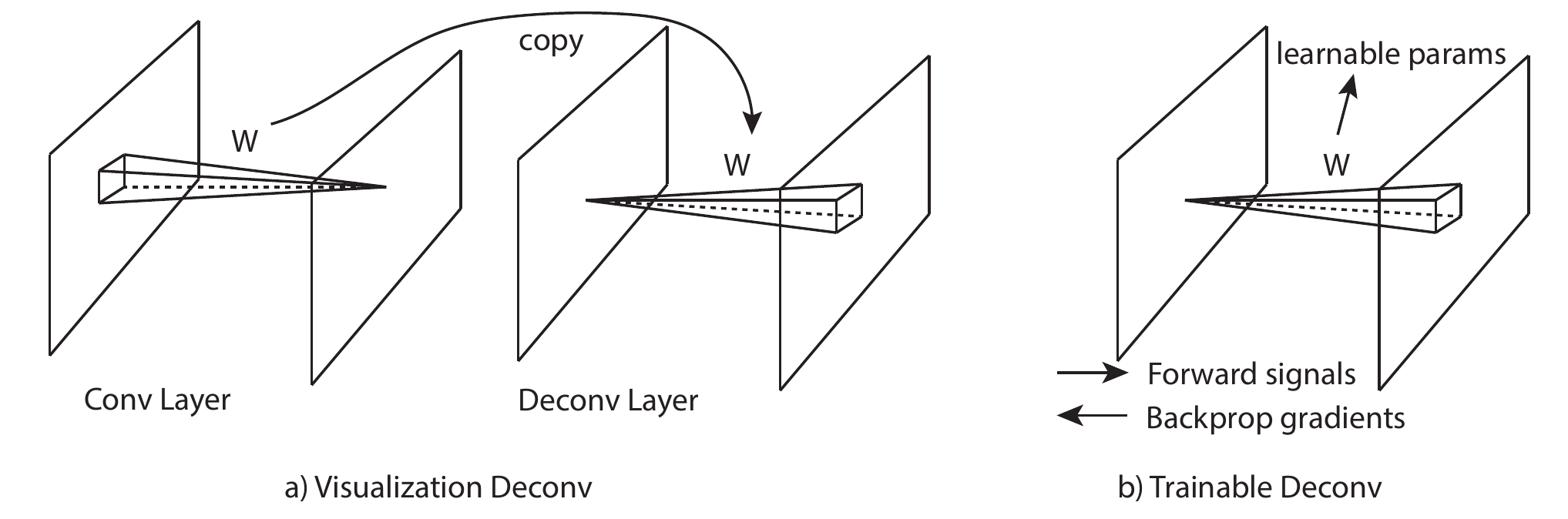}
\end{center}
\vspace{-12pt}
   \caption{{\bf Deconvolutional layers in ConvNets}. a) Visualization of the deconvolutional layer used in~\cite{ZeilerF14} where the filter weights are set to be equal to those of the pre-trained convolutional layer. b) Trainable deconvolutional layers~\cite{LongSD15} learn upsampling.}
\label{fig:deconv_layer}
\end{figure}

{\bf Deconvolution}. The concept of deconvolution was introduced by Zeiler and Fergus~\cite{ZeilerF14} to visualize the internal-layer filters of a 2D ConvNet. Because the objective of this prior work was merely filter visualization, there was no learning involved in the deconvolution layers and the weights were simply set to be equal to the transpose of the corresponding pre-trained convolution layers. Instead, Long~\emph{et al.}~\cite{LongSD15}  introduced the idea of deconvolution as a trainable layer in 2D ConvNets with applications to image semantic segmentation. As shown in Figure~\ref{fig:deconv_layer}, a filter of a trainable deconvolution layer acts as a learnable local upsampling unit. In convolution, input signals are convolved by the kernel filter and one value is placed on the output plane. Conversely, deconvolution takes one value from the input, multiples the value by the weights in the filter, and place the result in the output channel. Thus, if the 2D filter has size $s \times s$, it generates a $s \times s$ output matrix for each pixel input. The output matrices can be stored either overlapping or not overlapping in the output channel. If not overlapping, then deconvolution with a $s \times s$ filter would upsample the input by a factor $s$ in both dimensions. When the output matrices overlap, their contributions in the overlap are summed up. The amount of output overlap depends on the output {\em stride}. If the output stride is bigger than $1$, then the deconvolution layer produces an outputs with size larger than the input, thus acts as an upsampler. 

In our architecture, we use {\em 3D} deconvolutional layers, instead of 2D deconvolutional layers. This means that the filters are deconvolved spatio-temporally, instead of only spatially as in 2D ConvNets. 

\begin{figure*}[t]
\begin{center}
   \includegraphics[width=0.9\linewidth]{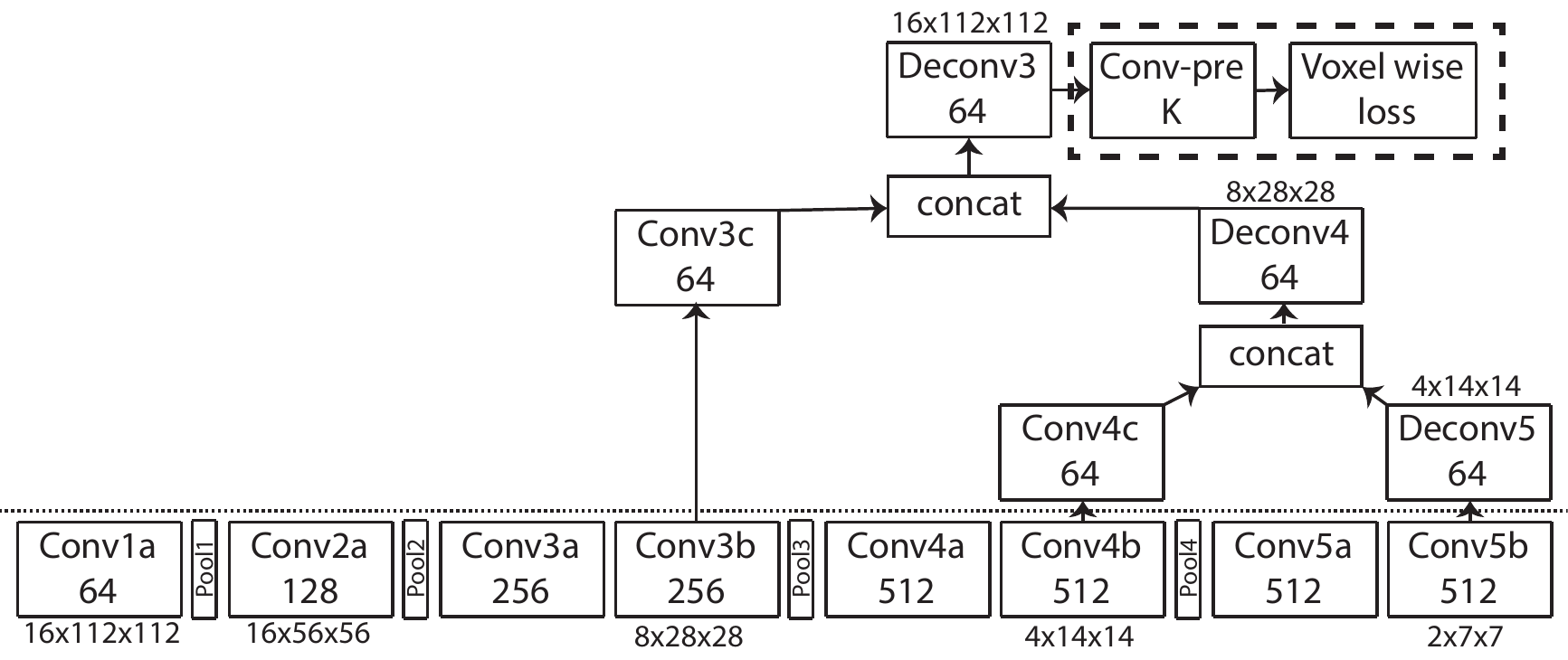}
\end{center}
\vspace{-12pt}
   \caption{{\bf V2V Architecture for Voxel Prediction}. The lower part (below dashed line) consists of layers from C3D~\cite{TranBFTP15}. Connected to these layers we have three 3D convolution layers: \texttt{Conv3c},\texttt{Conv4c},\texttt{Conv-pre} use filters of size $3 \times 3 \times 3$ with stride $1 \times 1 \times 1$. Both \texttt{Deconv5} and \texttt{Deconv4} are deconvolutional layers employing kernels of size $4 \times 4 \times 4$ with output stride of $2 \times 2 \times 2$. \texttt{Deconv3} has kernel size $8 \times 4 \times 4$ and output stride of $4 \times 2 \times 2$. The numbers inside the boxes represent the number of learning filters in that layer, while the numbers near the boxes (above or below) represent the size of output signals produced by that layer. The part inside the thick-dashed box is application-dependent.}
\label{fig:arch}
\end{figure*}

{\bf Architecture for voxel prediction}. Our architecture (which we name V2V, for voxel-to-voxel) is adapted from the C3D network described in~\cite{TranBFTP15}, which has shown good performance for different video recognition tasks. In order to apply it to voxel-prediction problems, we simply add 3D deconvolutional layers to the C3D network. Note that C3D operates by splitting the input video into clips of 16 frames each and perform prediction separately for each clip. Thus, our V2V model also takes as input a clip of 16 frames and then outputs voxel labels for the 16 input frames. Figure~\ref{fig:arch} illustrates our V2V architecture for voxel prediction. The lower part contains layers from C3D, while the upper part has {\em three} 3D convolutional layers, {\em three} 3D deconvolutional layers, {\em two} concatenation layers, and {\em one} loss layer. All three convolutional layers (\texttt{Conv3c},\texttt{Conv4c}, and \texttt{Conv-pre}) use filters of size $3 \times 3 \times 3$ with stride $1 \times 1 \times 1$ and padding $1 \times 1 \times 1$. \texttt{Conv3c} and \texttt{Conv4c} act as feature-map reducers, while \texttt{Conv-pre} acts as a prediction layer. \texttt{Deconv5} and \texttt{Deconv4} use filters of size $4 \times 4 \times 4$ with output stride $2 \times 2 \times 2$ and padding $1 \times 1 \times 1$. The \texttt{Deconv3} layer uses kernels of size $8 \times 4 \times 4$, an output stride of $4 \times 2 \times 2$, and padding $2 \times 1 \times 1$. Note that the number written inside the box of each layer in the Figure indicates the number of filters (e.g., 64 for \texttt{Deconv3}). The voxel-wise loss layer and \texttt{Conv-pre} layer are application-dependent and will be described separately for each of the applications considered in this paper. 
Since V2V shares the bottom layers with C3D, we have the option to either fine-tuning these layers starting from the C3D weights, or learning the weights from scratch.  We will report results for both options in our experiments.

\section{Application I: Video Semantic Segmentation}

{\bf Dataset}. Our experiments for video semantic segmentation are carried out on the GATECH  dataset~\cite{RazaGE13}, which comes with a public training/test split. The training set contains $63$ videos while the test set has $38$ sequences. There are 8 semantic classes: sky, ground, solid (mainly buildings), porous (mainly trees), cars, humans, vertical mix, and main mix.

{\bf Training}. Similarly to C3D, we down-scale the video frames to size $128 \times 171$. Because the dataset is quite small, we split each training video into all possible clips of length 16 (thus, we take overlapping clips with stride 1). For testing, we perform prediction on all non-overlapping clips of the video (stride equal to 16). We use the V2V architecture described in section~\ref{sec:formulation} with $K=8$ prediction channels, corresponding to the $8$ semantic classes. We use a voxel-wise softmax for the loss layer. We fine-tune the full V2V network initialized from C3D, using randomly initialized weights for the new layers. The learning rate is set initially to $10^{-4}$, and it is divided by $10$ every 30K iterations. The size of each mini-batch is $1$. Fine-tuning is stopped at 100K iterations, approximately $9$ epochs.

{\bf Baselines}. We compare our V2V model with several baselines to gain better insights about our method. The first set of baselines are based on bilinear upsampling. The purpose of these baselines is to understand the benefits of our 3D deconvolution layers compared to simple upsampling. Instead of using V2V with deconvolution layers, we use only C3D up to \texttt{Conv5b}, we then add a prediction layer (analogous to \texttt{Conv-pre}). Because the prediction made at \texttt{Conv5b} has size $2 \times 7 \times 7$, we apply a bilinear upsampling to produce a prediction of the same size as the input. We call this baseline \text{Conv5b-up}. We include two other baselines, namely, \text{Conv4b-up} and \text{Conv3b-up}, corresponding to adding a prediction layer and an upsampling layer at \text{Conv4b} and \text{Conv3b}, respectively. Besides these upsampling baselines, we also compare our fine-tuned V2V model with the V2V architecture trained from scratch on GATECH, which we call  V2V-0. We also trained a 2D version of V2V, namely 2D-V2V. The model 2D-V2V has the same architecture as V2V except that all 3D convolutional layers, 3D pooling layers, and 3D deconvolutional layers are replaced with 2D convolutional layers, 2D pooling layers, and 2D deconvolutional layers, respectively. As we do not a have pre-trained model of 2D-V2V, we train 2D-V2V from scratch on GATECH.

\begin{figure*}[t]
\begin{center}
   \includegraphics[width=0.9\linewidth]{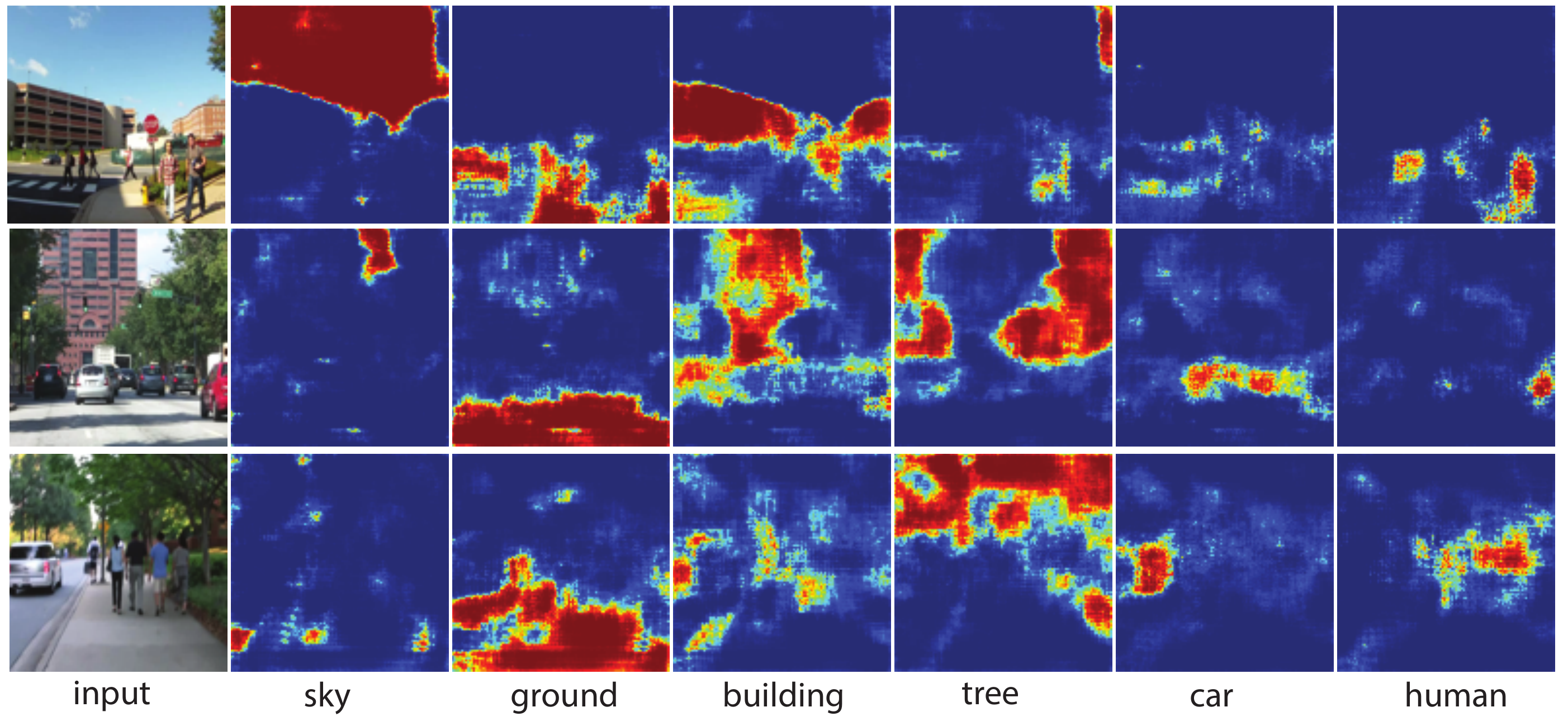}
\end{center}
\vspace{-12pt}
   \caption{{\bf Video Semantic Segmentation Results on GATECH}. The softmax prediction heat maps produced by V2V for different classes together with input frames. The last two classes are omitted due to their small populations. Best viewed in color.}
\label{fig:video_seg}
\end{figure*}

\begin{figure*}[t]
\begin{center}
   \includegraphics[width=0.9\linewidth]{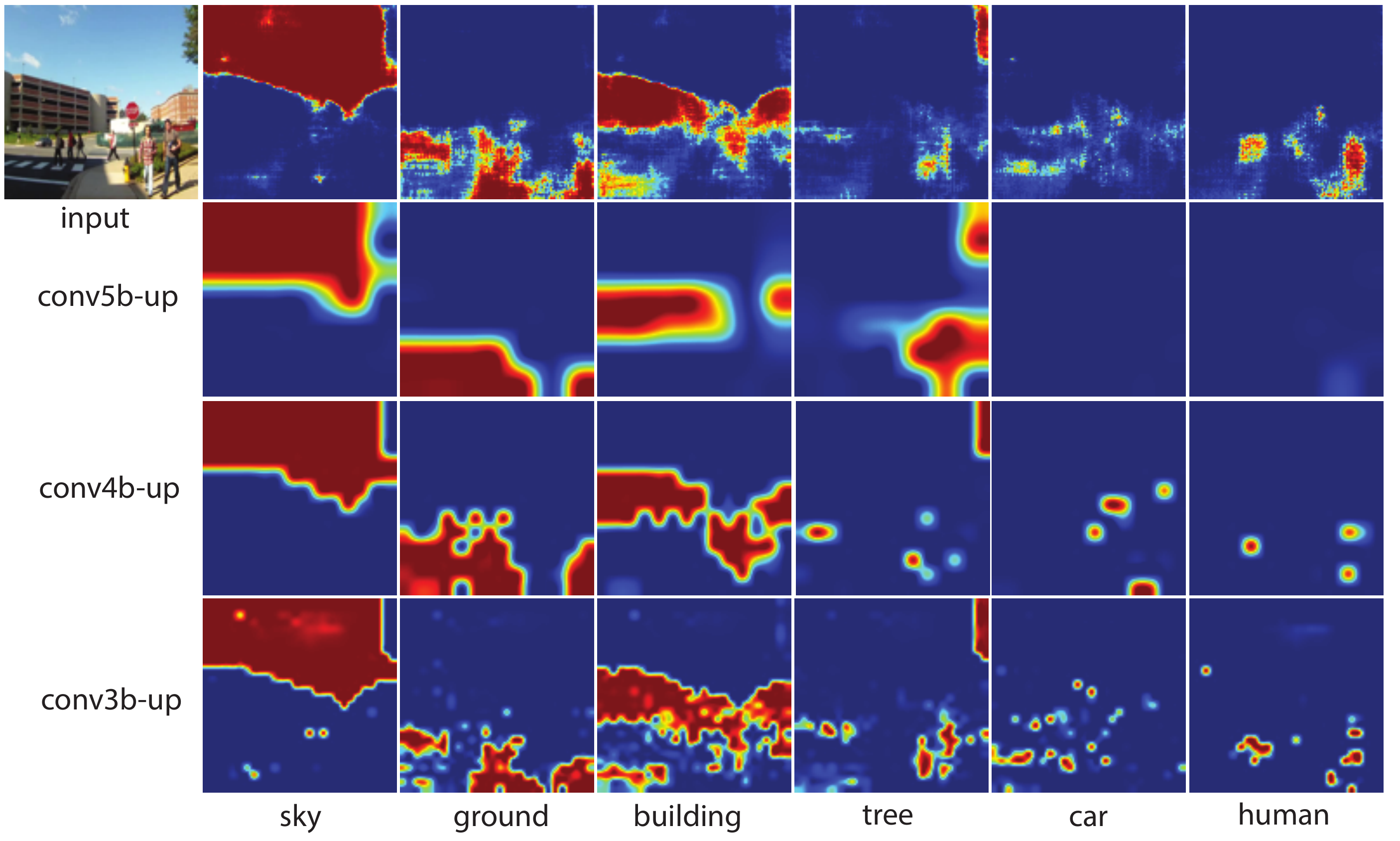}
\end{center}
\vspace{-12pt}
   \caption{{\bf V2V (top row) compared with upsampling baselines (rows 2-4)}. V2V consistently outperforms all bi-linear upsampling baselines. Conv5b-Up provides fairly accurate prediction, but over-smoothed due to the high upsampling factor. Conversely, Conv3b-Up yields finer predictions, but more noisy because it uses less deep features. V2V gives by far the best tradeoff as it has access to deep features and it learns the upsampling filters.}
\label{fig:video_seg_baseline}
\end{figure*}

{\bf Results}. Figure~\ref{fig:video_seg} visualizes some qualitative results of semantic segmentation using V2V on GATECH. Table~\ref{tab:video_segmenation_results} presents the semantic segmentation accuracy on GATECH of V2V compared with all of the baselines. 2D-V2V, trained from scratch on GATECH, obtains $55.7\%$ which is $11\%$ below V2V-0. This result underscores the advantages of 3D convolution and 3D deconvolution over their 2D counterparts. Note also that V2V-0 is $9.3\%$ below V2V. This predictably confirms the benefit of large-scale pre-training before fine-tuning. Finally, V2V also outperforms all bilinear upsampling baselines showing the advantages of using deconvolution over traditional upsampling. More qualitative comparisons of V2V with upsampling baselines are presented in Figure~\ref{fig:video_seg_baseline}. Here we can see that Conv5b-Up yields fairly accurate predictions but over-smoothed due to its big upsampling rate. On the other extreme, Conv3b-up produces finer predictions thanks to the lower upsampling rate, but its segments are noisy and fragmented because it relies on feature maps at layer 3, thus less deep and less complex than those used by Conv5b-Up.

\begin{table}
\begin{center}

\begin{tabular}{|l|c|c|}
\hline
Method & Train & Accuracy ($\%$)\\
\hline
 2D-V2V & from scratch & 55.7\\ 
 V2V-0 & from scratch & 66.7\\
 Conv3b+Up & fine-tune & 69.7\\
 Conv4b+Up & fine-tune & 72.7\\
 Conv5b+Up & fine-tune & 72.1\\
 {\bf V2V} & fine-tune & {\bf 76.0} \\
\hline
\end{tabular}
\end{center}
\vspace{-6pt}
\caption{{\bf Semantic segmentation accuracy on GATECH}. V2V consistently outperforms all baselines showing the good benefits of using V2V with 3D convolution/deconvolution compared to 2D convolution/deconvolution or bilinear upsampling. }
\label{tab:video_segmenation_results}
\end{table}

\section{Application II: Optical Flow Estimation}
{\bf Dataset}. Since there is no large-scale video dataset available with optical flow ground truth, we fabricate our training data by applying an existing optical flow method on unlabeled video. Specifically, we use the OpenCV GPU implementation of Brox's method~\cite{BroxM11} to generate semi-truth data on both UCF101~\cite{UCF101} (public test split 1) and MPI-Sintel~\cite{Sintel} (training set).  

{\bf Training}. We use the same V2V architecture with the number of channels at prediction layer set to $K=2$. On both horizontal and vertical motion components, we use the Huber loss for regression as it works well with noisy data and outliers. Formally, this is given by
\begin{equation}
H(x) = \\
\left\{
  \begin{array}{ll}
    \frac{1}{2}x^2, & |x| \leq 1\\
    |x|, & \text{otherwise}.
  \end{array}
\right.
\label{equ:huber}
\end{equation}   
To avoid numerical issues, the optical flow values are divided by a constant ($\alpha = 15$) so that most values fall in the range of $[-1,1]$. We note that larger optical flows are still handled by the Huber loss. The V2V network takes as input clips of size $3 \times 16 \times 112 \times 112$ and produces clip outputs of size $2 \times 16 \times 112 \times 112$. The network is trained from scratch on UCF101 (using non-overlapping clips from each video) with a mini-batch size of $1$. The initial learning rate is set to $10^{-8}$ and it is divided by $10$ every 200K iterations (about 2 epochs). Training is stopped at 800K iterations. We note that, at inference time, we need to scale the predictions by $\alpha=15$ to convert them back into the correct optical flow range.

\begin{figure*}[t]
\begin{center}
   \includegraphics[width=0.9\linewidth]{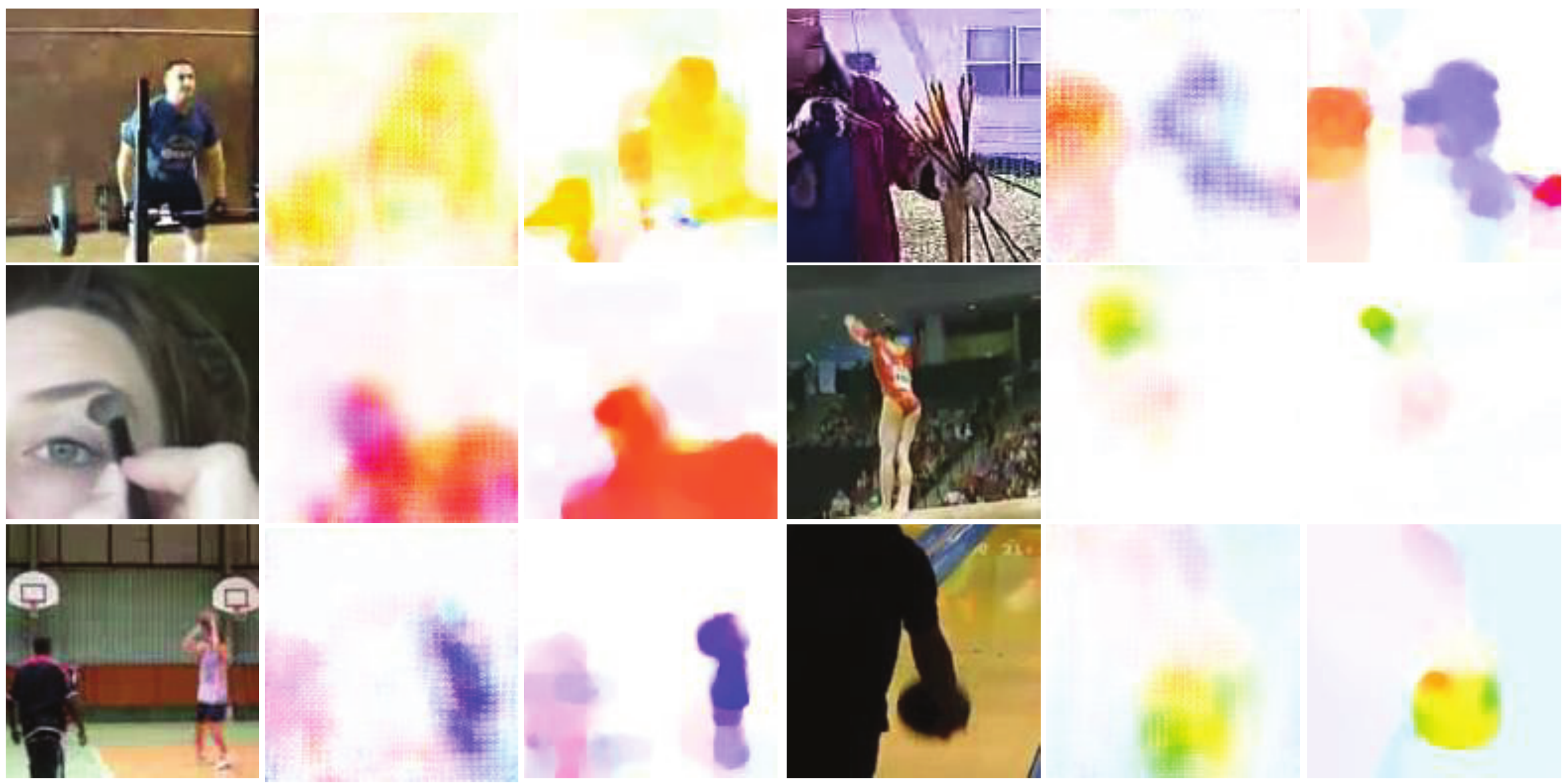}
\end{center}
\vspace{-12pt}
   \caption{{\bf Optical flow estimation on UCF101}. The output of V2V is qualitatively compared with Brox's optical flow for 6 sample clips from the UCF101 {\em test} split. For each example we show (from left to right): an input frame, V2V's predicted optical flow, and Brox's motion. Note that Brox's method is used to generate semi-truth data for training V2V. We see that on test videos V2V is able to predict flow of similar quality as that produced by Brox's algorithm. Best viewed in color.}
\label{fig:flow_ucf101}
\end{figure*}

\begin{table}
{\small
\begin{center}
\begin{tabular}{|l|c|c|}
\hline
Method & Brox & V2V-Flow \\
\hline
Run-time (hours) & 202.6 & 2.8 \\
FPS & 1.3 & 91.6 \\
x Slower & 70.5 & 1 \\
\hline
\end{tabular}
\end{center}}
\vspace{-6pt}
\caption{{\bf Runtime comparison}. The first row reports the total runtime (including I/O) to extract optical flow using V2V-Flow and Brox's method~\cite{BroxM11} for the entire UCF101 test split 1. V2V-Flow is $70$x faster than Brox's method, besides being slightly more accurate (see Table 2 of main paper).}
\label{tab:runtime}
\vspace{-12pt}
\end{table}

{\bf Results}. Figure~\ref{fig:flow_ucf101} visualizes optical flow predicted by our V2V method and compares it with that computed by Brox's method for a few sample clips taken from the test split of UCF101. The V2V end point error (EPE) on the UCF101 test split 1 (treating Brox's optical flow as ground truth) is only $1.24$. To better understand the performance of the learned V2V network, we further evaluate its performance on the training set of the MPI-Sintel dataset~\cite{Sintel}, which comes with ground truth data. This ground truth data is unbiased and allows us to assess performance independently from the accuracy of Brox's flow. Table~\ref{tab:sintel} shows the EPE error obtained with two variants of our model: V2V stands for our network learned on the UCF101 Brox's flow, while finetuned-V2V denotes our model after fine-tuning V2V on Sintel ground truth data using 3-fold cross validation. The table also contains the best method on Sintel which is better than V2V by a good margin. Even though V2V is not state of the art, the results are very interesting: both V2V and finetuned-V2V perform better than their ``teacher'', the optical flow method that is used to generate the semi-truth training data. While the improvement is slim, it is important to highlight that V2V is {\em much faster} than Brox's algorithm ($70$x faster, see Table~\ref{tab:runtime}). Thus, this experiment shows that the V2V network can be employed to learn efficient implementations of complex, hand-tuned voxel-prediction models.

Table~\ref{tab:runtime} presents the detailed runtime comparison between V2V-Flow and Brox's method~\cite{BroxM11}. We use the GPU implementation of Brox's method provided in OpenCV. Table~\ref{tab:runtime} reports the runtime (including I/O) to extract optical flow for the whole UCF101 test split $1$ by the two methods using a NVIDIA Tesla K40. V2V-Flow is $70$x faster than Brox's method. It can run at $91$ fps while Brox's method operates at less than $2$ fps.

\begin{table}
{\small
\begin{center}
\begin{tabular}{|l|c|c|c|c|}
\hline
Method & Brox & V2V & finetuned-V2V & FlowFields~\cite{FlowFields} \\
\hline
EPE & 8.89 & 8.86 & 8.38 & {\bf 5.81} \\
\hline
\end{tabular}
\end{center}}
\vspace{-6pt}
\caption{{\bf Optical flow results on Sintel}. V2V denotes our network learned from the UCF101 optical flow computed with Brox's method. The finetuned-V2V network is obtained by fine-tuning V2V on Sintel (test accuracy is measured in this case using 3-fold cross validation). Both versions of our network perform slightly better than Brox's algorithm and they allow computation of optical flow with a runtime speedup of 20 times compared to Brox's software.}
\label{tab:sintel}
\vspace{-12pt}
\end{table}

\begin{figure*}[t]
\begin{center}
   \includegraphics[width=0.9\linewidth]{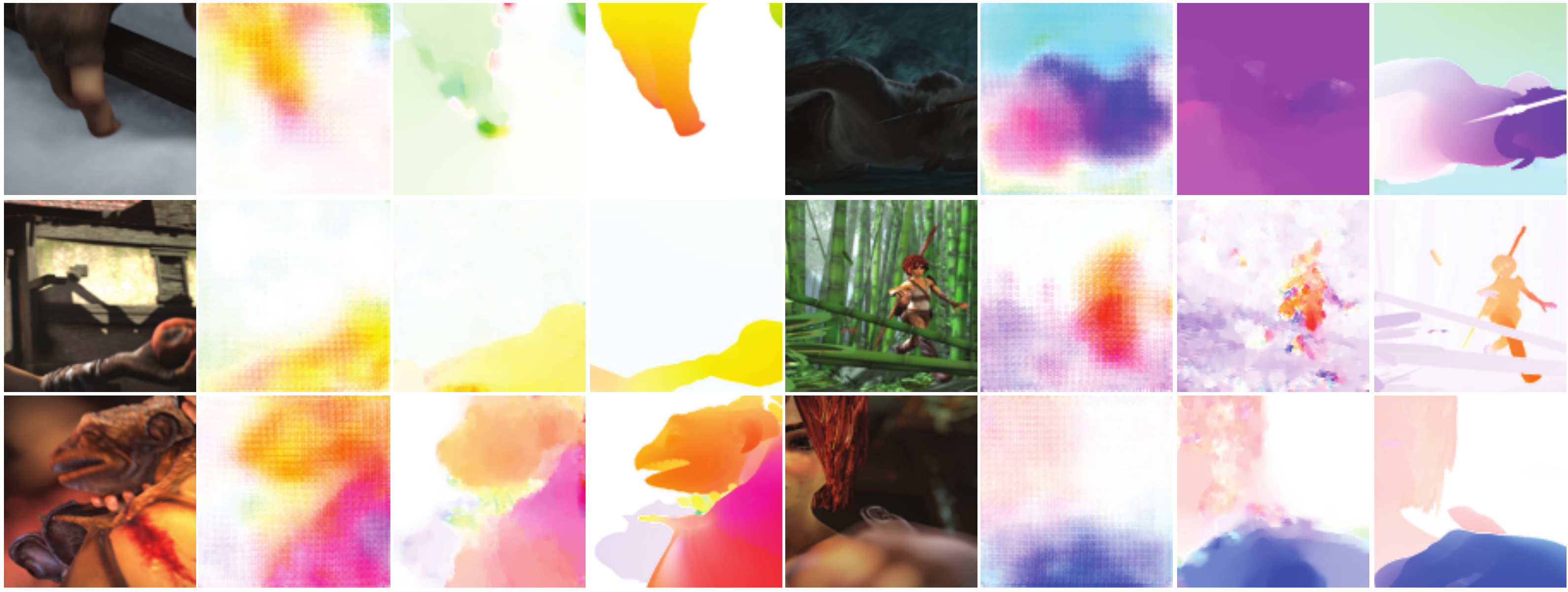}
\end{center}
\vspace{-12pt}
   \caption{Visualizations of optical flow computed by the V2V network (trained on UCF101 without finetuning) for a few sample Sintel clips. For each example we show: input frame, V2V's predicted optical flow, Brox's flow, and ground truth. Best viewed in color.}
\label{fig:flow_sintel}
\vspace{-12pt}
\end{figure*}

\begin{figure*}[t]
\begin{center}
   \includegraphics[width=0.9\linewidth]{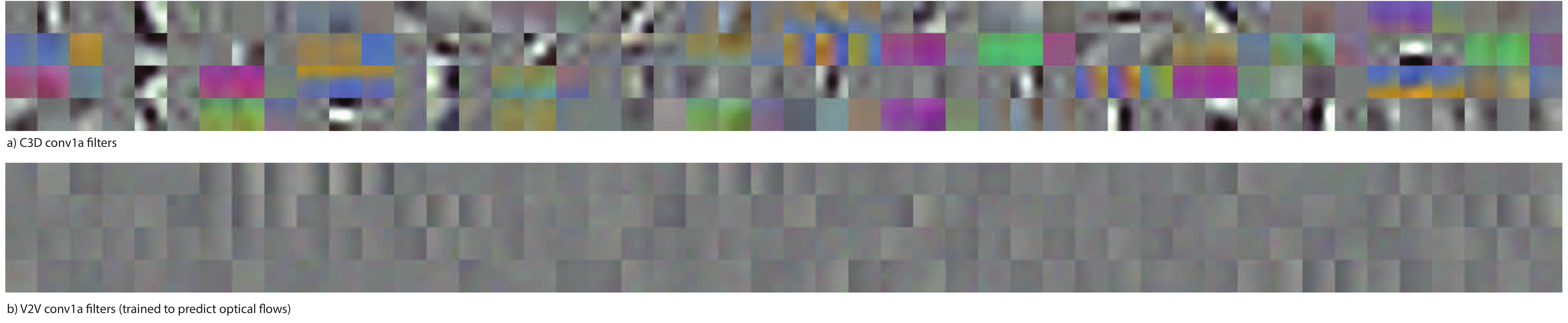}
\end{center}
\vspace{-12pt}
   \caption{Visualization of \texttt{Conv1a} filters learned by C3D (top) and V2V (bottom). Note that C3D is trained to recognize actions (on Sport1M), while V2V is optimized to estimate optical flow (on UCF101). Each set shows the $64$ learned filters at the \texttt{Conv1a} layer. Three consecutive square images on each row represent one filter (as kernel size is $3 \times 3 \times 3$). Each square image is upscaled to $30 \times 30$ pixels for better visualization. Best viewed in color. \emph{GIF animation of these filters will be provided in the project website}.}
\label{fig:conv1a_comparison}
\end{figure*}

{\bf Observation}. Unlike the case of video semantic segmentation application where V2V could be effectively fine-tuned from the initial C3D network, we empirically discovered that fine-tuning from C3D does not work for the case of optical flow estimation as in this case the training consistently converges to a bad local minimum. We further investigated this phenomenon by visualizing the learned filers of the first few convolutional layers for both the original C3D as well as the V2V learned from scratch on Brox's flow. The results are visible in Fig.~\ref{fig:conv1a_comparison}. We see that the filters of the two networks look completely different. This is understandable, as C3D is trained to complete a high-level vision task, e.g. classifying sports. Thus the network learns a set of discriminative filters at the early layers. Some of these filters capture texture, some focus on discriminative motion patterns, while others respond to particular appearance or color cues. Instead, V2V is trained to perform a low-level vision task, e.g. predict motion directions. The Figure shows that the V2V filters are insensitive to color and texture as they focus exclusively on motion estimation. This explains why the pre-trained C3D model is a bad initialization to learn V2V for optical flow, but it is instead a good initialization for training V2V on semantic segmentation. 

\section{Application III: Video Coloring}
{\bf Setup and Training}. In this experiment we use UCF101 again in order to learn to color videos. We use the public training/test split 1 for the training and testing of our model. In this study we generate training data by converting the color videos to grayscale. V2V is fed with $C=1$ input grayscale channel and it is optimized to predict the $K=3$ ground truth original color channels. For this application we use the L2 regression loss as colors have no outliers. We use mini-batches of size $1$. The learning rate is set initially to $10^{-8}$ and it is divided by $10$ every 200K iterations. The training is stopped at 600K iterations. Similarly to the case of semantic segmentation, we compare our V2V with its 2D version baseline, 2D-V2V, both optimized on the same training set. Both models were learned from scratch.

\begin{figure*}[t]
\begin{center}
   \includegraphics[width=0.9\linewidth]{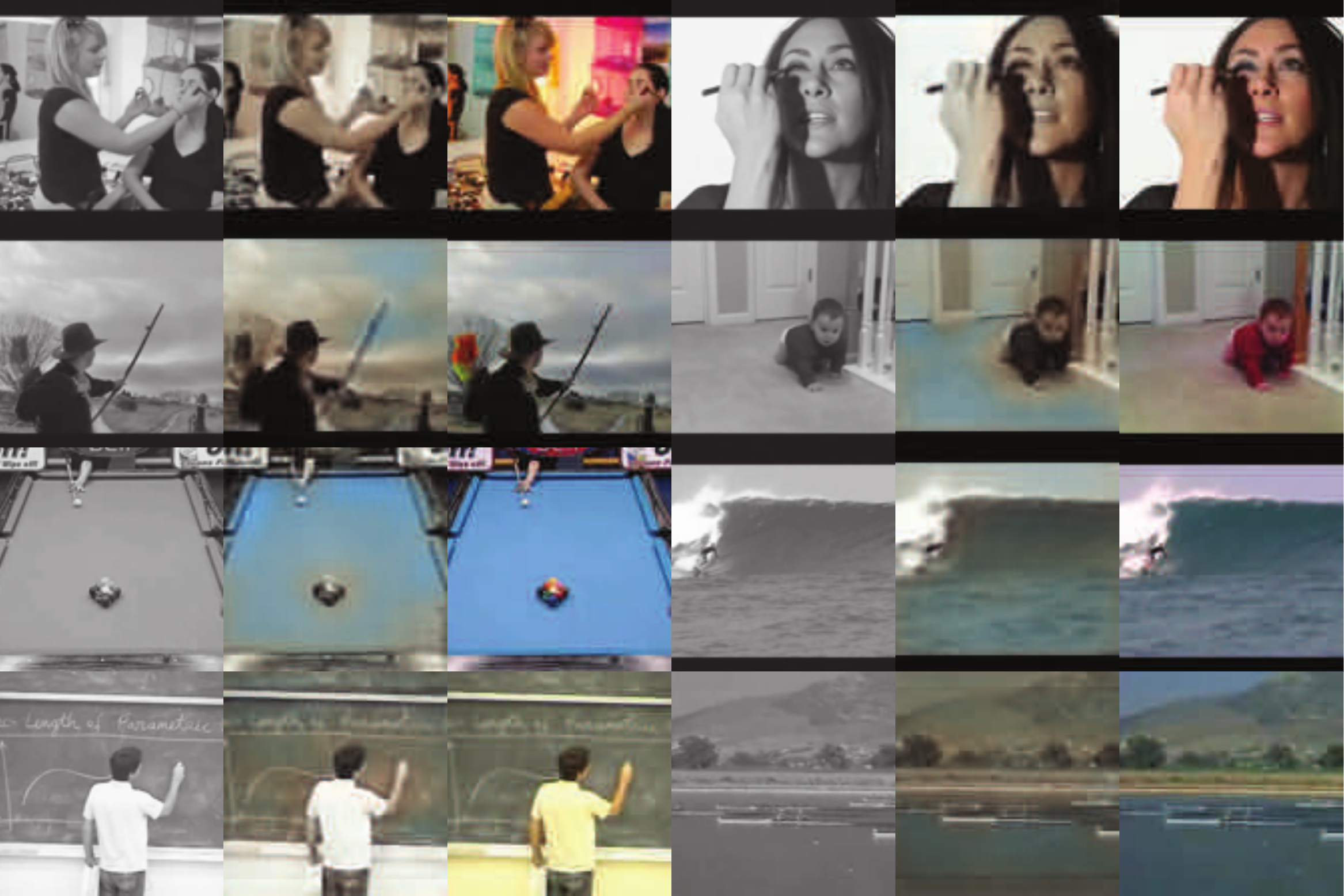}
\end{center}
\vspace{-12pt}
   \caption{{\bf Examples of video coloring with V2V on the test set of UCF101}. For each example we show (from left to right): a gray-scale input frame, the output frame colored by V2V, and the ground truth color frame. The V2V model is able to predict ``common sense'' colors such as the color of human skin, sky, woody furniture, river, sea, and mountain. Best viewed in color.}
\label{fig:video_coloring}
\end{figure*}

We note that video coloring is challenging and ill-posed because there are some objects (e.g., clothes) that can be colored with any valid color. A reasonable expectation is that the coloring algorithm should learn to color correctly objects that typically occur only in one color. For example, the sky is usually blue (not always but often) and the grass is typically green. Thus, the model should learn to predict well the colors of such objects.

{\bf Results}. To assess performance, we use as metric the average Euclidean distance between the predicted color and the true color. Here each voxel color is represented in $(r,g,b)$ and $r,g,b \in [0,1]$. V2V has an average distance error (ADE) of $0.1375$ whereas the 2D baseline has an ADE of $0.1495$. Figure~\ref{fig:video_coloring} presents some qualitative results of V2V on predicting voxel colors. It is interesting to see that the algorithm learns ``common sense'' colors such as the color of skin, sky, trees, river, sea, mountains, wood furniture, and the billiard table. For objects whose color is ambiguous, V2V applies very little coloring, leaving them almost in the original grayscale form. One can imagine extending V2V to have sparse inputs of color to make the problem well-posed for objects that can occur in various colors.

\section{Conclusions}

We have presented V2V, a novel architecture for voxel to voxel prediction using 3D convolutional networks. The proposed approach can be trained end to end from raw video input to predict target voxel labels without the need to preprocess or post-process the data. We have shown that the same architecture trained on three distinct application domains delivers competitive results on each of them. In the course of our experiments we have discovered that fine-tuning pre-trained models does not always help: for the case of optical flow estimation, learning from scratch is beneficial over fine-tuning from an action recognition model. We have also demonstrated that in absence of large-scale supervised data, V2V can be trained to reproduce the output of an existing hand-constructed voxel prediction model. Quite surprisingly, in our study the resulting learned model has accuracy superior (albeit only slightly) to its ``teacher'' method. We believe that bootstrapping the learning from an existing model can be an interesting avenue for future work and can be a successful strategy to learn efficient implementation of computationally expensive algorithm, such as in our case where V2V predicts optical flow with a $70$x speedup over the original optical flow method that was used to generate training data. While we purposely avoided specializing the network to each task in order to emphasize the general applicability of  the approach, we believe that further improvements can be obtained from more thorough architecture search.

{\bf Acknowledgment}: we would like to thank colleagues at Facebook AI Research and Dartmouth Vision and Learning Group for valuable feedback and discussions.

{\small
\bibliographystyle{ieee}
\bibliography{ieeedu_ref}

\begin{thebibliography}{10}\itemsep=-1pt

\bibitem{FlowFields}
C.~Bailer, B.~Taetz, and D.~Stricker.
\newblock Flow fields: Dense correspondence fields for highly accurate large
  displacement optical flow estimation.
\newblock In {\em ICCV}, 2015.

\bibitem{Bertasius15}
G.~Bertasius, J.~Shi, and L.~Torresani.
\newblock Deepedge: A multi-scale bifurcated deep network for top-down contour
  detection.
\newblock In {\em CVPR}, 2015.

\bibitem{Irani05}
M.~Blank, L.~Gorelick, E.~Shechtman, M.~Irani, and R.~Basri.
\newblock Actions as space-time shapes.
\newblock In {\em ICCV}, pages 1395--1402, 2005.

\bibitem{BroxM11}
T.~Brox and J.~Malik.
\newblock Large displacement optical flow: Descriptor matching in variational
  motion estimation.
\newblock {\em IEEE TPAMI}, 33(3):500--513, 2011.

\bibitem{Sintel}
D.~Butler, J.~Wulff, G.~Stanley, and M.~Black.
\newblock A naturalistic open source movie for optical flow evaluation.
\newblock In {\em CVPR}, 2012.

\bibitem{LCao10}
L.~Cao, Z.~Liu, and T.~Huang.
\newblock Cross-dataset action detection.
\newblock In {\em Proc. IEEE Conference on Computer Vision and Pattern
  Recognition}, 2010.

\bibitem{Donahue13}
J.~Donahue, Y.~Jia, O.~Vinyals, J.~Hoffman, N.~Zhang, E.~Tzeng, and T.~Darrell.
\newblock Decaf: {A} deep convolutional activation feature for generic visual
  recognition.
\newblock In {\em ICML}, 2013.

\bibitem{Efros03}
A.~Efros, A.~Berg, G.~Mori, and J.~Malik.
\newblock Recognizing action at a distance.
\newblock In {\em Proc. International Conference on Computer Vision}, pages
  726--733, 2003.

\bibitem{EigenNIPS14}
D.~Eigen, C.~Puhrsch, and R.~Fergus.
\newblock Depth map prediction from a single image using a multi-scale deep
  network.
\newblock In {\em NIPS}, 2014.

\bibitem{FlowNet}
P.~Fischer, A.~Dosovitskiy, E.~Ilg, P.~H{\"{a}}usser, C.~Hazirbas, V.~Golkov,
  P.~Smagt, D.~Cremers, and T.~Brox.
\newblock Flownet: Learning optical flow with convolutional networks.
\newblock In {\em ICCV}, 2015.

\bibitem{girshick13}
R.~Girshick, J.~Donahue, T.~Darrell, and J.~Malik.
\newblock Rich feature hierarchies for accurate object detection and semantic
  segmentation.
\newblock {\em arXiv preprint arXiv:1311.2524}, 2013.

\bibitem{Ikizler08}
N.~Ikizler and D.~Forsyth.
\newblock Searching for complex human activities with no visual examples.
\newblock {\em International Journal of Computer Vision}, 80(3):337--357, 2008.

\bibitem{Jhuang07}
H.~Jhuang, T.~Serre, L.~Wolf, and T.~Poggio.
\newblock A biological inspired system for human action classification.
\newblock In {\em Proc. International Conference on Computer Vision}, 2007.

\bibitem{Karpathy14}
A.~Karpathy, G.~Toderici, S.~Shetty, T.~Leung, R.~Sukthankar, and L.~Fei-Fei.
\newblock Large-scale video classification with convolutional neural networks.
\newblock In {\em CVPR}, 2014.

\bibitem{Krizhevsky12}
A.~Krizhevsky, I.~Sutskever, and G.~Hinton.
\newblock Imagenet classification with deep convolutional neural networks.
\newblock In {\em NIPS}, 2012.

\bibitem{Laptev03}
I.~Laptev and T.~Lindeberg.
\newblock Space-time interest points.
\newblock In {\em ICCV}, 2003.

\bibitem{LongSD15}
J.~Long, E.~Shelhamer, and T.~Darrell.
\newblock Fully convolutional networks for semantic segmentation.
\newblock In {\em CVPR}, 2015.

\bibitem{Ng15}
J.~Ng, M.~Hausknecht, S.~Vijayanarasimhan, O.~Vinyals, R.~Monga, and
  G.~Toderici.
\newblock Beyond short snippets: Deep networks for video classification.
\newblock In {\em CVPR}, 2015.

\bibitem{FeiFei07}
J.~Niebles and L.~Fei-Fei.
\newblock A hierarchical model of shape and appearance for human action
  classification.
\newblock In {\em Proc. IEEE Conference on Computer Vision and Pattern
  Recognition}, pages 1--8, 2007.

\bibitem{RazaGE13}
S.~H. Raza, M.~Grundmann, and I.~Essa.
\newblock Geometric context from video.
\newblock In {\em CVPR}, 2013.

\bibitem{ActionBank}
S.~Sadanand and J.~Corso.
\newblock Action bank: A high-level representation of activity in video.
\newblock In {\em CVPR}, 2012.

\bibitem{HJSeo09}
H.~Seo and P.~Milanfar.
\newblock Detection of human actions from a single example.
\newblock In {\em Proc. International Conference on Computer Vision}, 2009.

\bibitem{overfeat}
P.~Sermanet, D.~Eigen, X.~Zhang, M.~Mathieu, R.~Fergus, and Y.~LeCun.
\newblock Overfeat: Integrated recognition, localization and detection using
  convolutional networks.
\newblock In {\em ICLR}, 2014.

\bibitem{SimonyanZ14}
K.~Simonyan and A.~Zisserman.
\newblock Two-stream convolutional networks for action recognition in videos.
\newblock In {\em NIPS}, 2014.

\bibitem{SimonyanZ14a}
K.~Simonyan and A.~Zisserman.
\newblock Very deep convolutional networks for large-scale image recognition.
\newblock In {\em ICLR}, 2015.

\bibitem{UCF101}
K.~Soomro, A.~R. Zamir, and M.~Shah.
\newblock {UCF101}: A dataset of 101 human action classes from videos in the
  wild.
\newblock In {\em CRCV-TR-12-01}, 2012.

\bibitem{Googlenet}
C.~Szegedy, W.~Liu, Y.~Jia, P.~Sermanet, S.~Reed, D.~Anguelov, D.~Erhan,
  V.~Vanhoucke, and A.~Rabinovich.
\newblock Going deeper with convolutions.
\newblock In {\em CVPR}, 2015.

\bibitem{TranBFTP15}
D.~Tran, L.~Bourdev, R.~Fergus, L.~Torresani, and M.~Paluri.
\newblock Learning spatiotemporal features with 3d convolutional networks.
\newblock In {\em ICCV}, 2015.

\bibitem{Wang2013IJCV}
H.~Wang, A.~Kl{\"a}ser, C.~Schmid, and C.-L. Liu.
\newblock {Dense trajectories and motion boundary descriptors for action
  recognition}.
\newblock {\em IJCV}, 103(1):60--79, 2013.

\bibitem{Yuan11}
J.~Yuan, Z.~Liu, and Y.~Wu.
\newblock Discriminative video pattern search for efficient action detection.
\newblock {\em IEEE Trans. on Pattern Analysis and Machine Intelligence}, 2011.

\bibitem{ZeilerF14}
M.~Zeiler and R.~Fergus.
\newblock Visualizing and understanding convolutional networks.
\newblock In {\em ECCV}, 2014.

\bibitem{Zhou2014}
B.~Zhou, A.~Lapedriza, J.~Xiao, A.~Torralba, and A.~Oliva.
\newblock Learning deep features for scene recognition using places database.
\newblock In {\em NIPS}, 2014.

\end{thebibliography}
}

\end{document}